\documentclass[runningheads,orivec]{llncs}

\usepackage[T1]{fontenc}
\usepackage{amsmath}
\usepackage{newtxtext}
\usepackage[varvw]{newtxmath}
\usepackage{booktabs}
\usepackage{tabularx}
\usepackage{graphicx}
\usepackage{float}
\usepackage{needspace}
\usepackage{microtype}
\usepackage{url}
\usepackage[hidelinks]{hyperref}
\AtBeginDocument{\renewcommand{\doi}[1]{\url{https://doi.org/#1}}}

\graphicspath{{figures/}}

\newcommand{\appendixfigure}[1]{%
    \includegraphics[width=\linewidth]{#1}%
}

\newcommand{\githubrepourl}{\url{https://github.com/har5h1l/affect_aif}}
\newcommand{\appendixsupplementarystatement}{%
  Simulation code, experiment configurations, analysis scripts, figure-generation code, and the frozen row-level result archive are available at~\githubrepourl.
}

\title{Partner-Specific Affective Precision in Social Active Inference}
\titlerunning{Partner-Specific Affective Precision}
\author{Harshil Shah\inst{1}\orcidID{\href{https://orcid.org/0009-0006-0145-9958}{0009-0006-0145-9958}}\thanks{Corresponding author: \email{28hshah@gmail.com}.} \and
Andrew Pashea\inst{2}\orcidID{\href{https://orcid.org/0009-0004-4061-6296}{0009-0004-4061-6296}}}
\authorrunning{H. Shah and A. Pashea}
\institute{Mission San Jose High School, Fremont, CA, USA
\and
Division of the Social Sciences, University of Chicago, Chicago, IL, USA}

\begin{document}
\raggedbottom
\maketitle

\begin{abstract}
\emergencystretch=2em
In multi-agent social settings, model reliability varies across relationships. Beyond inferring what others will do, an agent must calibrate how confidently those inferences should guide policy selection for each relationship. An agent may maintain a well-validated model of one partner, a fragile model of another, and a model under revision for a third; collapsing these into a single confidence estimate loses information relevant to policy selection. We therefore formalize affective precision as a relationship-specific metacognitive estimate of confidence in the current partner model. Each partner's behavioral evidence updates a local confidence estimate that modulates policy precision during selection, regulating how strongly current beliefs are expressed in policy rather than changing the content of those beliefs.

Simulations in a multi-partner graded trust game show that partner-local affective precision influences behavior primarily through policy commitment rather than direct improvement of partner-state inference. Because the mechanism tracks partner-response predictability rather than realized payoff, greater confidence produces sharper policy commitment without necessarily producing higher rewards. Under abrupt shifts in social behavior, confidence accumulated from previously reliable predictions can remain behaviorally active after the relationship changes, showing that confidence revision can lag behind social change. Finally, varying precision gain and priors produce distinct trust-calibration dynamics, showing how confidence accumulation and revision depend on model parameters. Together, these results show how relationship-specific affective precision can distinguish social prediction from social policy commitment.

\end{abstract}
\section{Introduction}

Social agents rarely act on a single, stable model of the social world; instead, they maintain different expectations for different partners. One partner may be familiar and reliable, another newly encountered and uncertain, and another predictable but adversarial. Repeated social exchange depends on learning such partner-specific regularities, including reciprocity, reputation, moral character, and social value~\cite{Behrens2008SocialLearning,Delgado2005MoralCharacter,KingCasas2005Trust}. In such settings, the question is not only what an agent believes about each partner, but how confidently those beliefs should guide policy inference.

Computational models of social decision-making can therefore be separated into three problems. The first is inferring a partner's behavioral disposition: whether their actions are cooperative, exploitative, reciprocal, or volatile. The second is modeling what they know or intend: the theory-of-mind problem of how another agent represents the world and plans action~\cite{FrithFrith2012,Hampton2008Mentalizing,Pitliya2025ToMActiveInference}. The third is estimating whether the resulting partner model is reliable enough to guide confident policy commitment~\cite{Schoeller2021TrustExtendedControl}. To address this third problem, an agent must decide not only what it believes about a partner, but how strongly to act on those beliefs. However, many formal models specify the content of social inference while treating confidence in those inferences as fixed, global, or implicit.

The active inference framework~\cite{Parr2022ActiveInference} provides a natural formalism for this problem because it links perception, belief updating, and policy selection through a common inferential process. In active inference, agents infer hidden states of the world and select policies by minimizing expected free energy. Moreover, policy inference is precision-weighted: policy precision controls how sharply an agent commits to its inferred policy distribution, making active inference a model not only of what an agent believes but how confidently those beliefs are expressed in policy~\cite{Friston2017ProcessTheory}.

Existing social active-inference models address complementary parts of the problem. Theory-of-mind models address how agents model what others know or intend through recursive belief structures and factorized generative models~\cite{Pitliya2025ToMActiveInference,RuizSerra2025Factorised}. Volatility and social-learning models address how prediction errors should change belief updating under uncertainty~\cite{Behrens2008SocialLearning,Diaconescu2017Hierarchical,Mathys2011HGF}. Social-learning work also shows that agents revise expectations from discrepancies between expected and observed behavior of others~\cite{joiner2017social}. Formal models of interpersonal inference have similarly applied active inference to repeated trust-game exchange, in which agents update beliefs about social partners and select policies under those beliefs~\cite{Moutoussis2014InterpersonalInference}. Related opponent-modeling and type-based approaches maintain hypotheses about other agents' behavioral strategies, goals, or beliefs in order to predict their behavior~\cite{AlbrechtStone2018ModellingAgents}. These approaches clarify how agents infer partners, model others' mental states, and revise beliefs under uncertainty. However, they do not by themselves explain how confidence in a social model should remain local to the relationship that generated the evidence and modulate policy commitment accordingly.

Affect provides a natural candidate mechanism for this confidence-calibration role. In psychology and neuroscience, affect is often described in terms of valence and arousal---the felt pull toward or away from situations and social commitments~\cite{Barrett2006EmotionParadox}. Free-energy and active-inference accounts extend this view by linking affective dynamics to prediction error, model evidence, emotional-state inference, and precision-weighted policy inference~\cite{Hesp2021DeeplyFeltAffect,joffily2013emotional,pattisapu2024free,smith2019simulating}. Prior affective-precision accounts make confidence endogenous by relating affect to subjective model fitness and policy precision~\cite{Hesp2021DeeplyFeltAffect}. On this view, affect can regulate how strongly current beliefs are expressed in policy, shaping the degree of commitment rather than the content of inference. In this work, \emph{affective precision} refers specifically to confidence in deploying a partner model during policy selection. It concerns how strongly partner beliefs guide policy rather than felt valence, emotion concepts, attachment, appraisal, or affective experience more broadly.

Related hierarchical active-inference work has also modeled metacognitive states that regulate precision within the generative hierarchy~\cite{SandvedSmith2021MentalAction}. Here, we focus specifically on making such confidence relationship-specific and linking it to the predictive reliability of individual partner models.

We therefore define partner-local affective precision as a relationship-specific confidence signal. For each partner \(k\), the focal agent maintains a posterior \(q(\beta_k)\) over an inverse policy-precision variable \(\beta_k\), such that lower \(\beta_k\) corresponds to greater confidence and sharper policy selection. Evidence from partner \(k\) updates only that partner's confidence estimate, and its posterior mean modulates policy precision for policies evaluated under the corresponding partner model. Unlike prior affective-precision accounts centered on policy-inference quantities~\cite{Hesp2021DeeplyFeltAffect}, we derive the affective signal from the predictive evidence assigned to the partner's observed response under the focal agent's current model. The signal therefore tracks whether the current model predicts this partner's behavior rather than whether the interaction produces a favorable payoff, preserving relationship-specific differences that a global confidence estimate would collapse.

We evaluate this mechanism in a repeated multi-partner graded trust game with partner-choice conditions, abrupt betrayal, and profile variation. The remainder of this paper is organized as follows: Section~\ref{sec:methods} specifies the multi-partner trust game, per-partner generative models, and the partner-local affective precision mechanism. Section~\ref{sec:results} reports simulation results and analyses. Section~\ref{sec:discussion} discusses implications, limitations, and directions for future work.
\section{Methods}
\label{sec:methods}

A focal active-inference agent plays a repeated multi-partner trust/investment game with four partners whose behavioral types and stances are hidden. The agent is implemented using the \texttt{pymdp} library for discrete active inference~\cite{DaCosta2020DiscreteSynthesis,Heins2022Pymdp}, which provides standard components for belief updating and policy selection. Partner-local affective precision is layered on top of separate per-partner generative models, allowing both evidence and policy precision to remain relationship-specific. The task generative process, focal-agent POMDP, affective-precision update, and simulation protocols are specified in Appendices~\ref{app:generative_process}--\ref{app:protocols}.

\subsection{Trust Game}
\label{sec:trust_game}

We evaluate the mechanism in a repeated multi-partner trust/investment game~\cite{Berg1995Trust}. The task has a Prisoner's-Dilemma-style social-dilemma structure: investing more can yield larger returns when a partner cooperates, but larger losses relative to the risk-free baseline when the partner defects. The task therefore requires the focal agent to solve two coupled problems: predicting how each partner is likely to respond, and deciding how strongly to commit to that prediction through investment.

Each interaction consists of partner assignment or selection, investment, and response. In partner-choice conditions, the focal agent first selects a partner \(k\); in conditions without explicit partner choice, the interaction partner is assigned according to the configured protocol. The agent then chooses a graded investment level \(a \in \mathcal{A}=\{0,1,2,3,4,5\}\), where \(0\) denotes no investment and \(5\) denotes maximum investment. The selected partner emits a response \(o_k^{\mathrm{act}} \in \{\mathrm{cooperate},\mathrm{defect}\}\), sampled from that partner's hidden behavioral state.

Each partner has two hidden state factors. The first is behavioral type, \(s_k^{\mathrm{type}} \in \{\mathrm{cooperator},\mathrm{reciprocator},\mathrm{exploiter},\mathrm{random}\}\). These labels are stylized abstractions of heterogeneous motives and reciprocity patterns observed in trust-game behavior~\cite{Espin2016HeterogeneousMotivesTrust,Smith2013ReciprocityTrustGame}. The second is stance toward the focal agent, \(s_k^{\mathrm{stance}} \in \{\mathrm{trusting},\mathrm{neutral},\mathrm{hostile}\}\). These states are not directly observed; the focal agent infers them from partner responses and payoff observations. Partners themselves are environment-side processes: they sample cooperate-or-defect responses from type- and stance-conditioned probabilities and undergo investment-dependent stance transitions, but do not perform variational inference or maintain affective precision estimates themselves. The present study thus centers on the focal agent's precision mechanism, whereas future work may extend this implementation to reciprocal active-inference partners.

With endowment \(E=10\) and return multiplier \(M=3\), when the partner cooperates, the invested amount is multiplied by \(M\) and split evenly, so the focal agent receives \(Ma/2\); when the partner defects, the focal agent loses the invested amount. Focal payoff is
\[
\pi(a,o_k^{\mathrm{act}})=
\begin{cases}
E-a+\frac{Ma}{2}, & o_k^{\mathrm{act}}=\mathrm{cooperate},\\
E-a, & o_k^{\mathrm{act}}=\mathrm{defect}.
\end{cases}
\]
Thus, cooperation yields \(\pi(a,\mathrm{cooperate})=10+0.5a\), whereas defection yields \(\pi(a,\mathrm{defect})=10-a\). The no-investment action \(a=0\) gives a risk-free payoff of \(10\), while higher investment increases both potential gains and potential loss.

Investment also affects future partner stance. Higher investment provides stronger cooperative evidence and shifts stance dynamics toward trust-building transitions; lower investment or withholding shifts dynamics toward trust-damaging transitions. Formally, the transition model interpolates between high-investment and low-investment endpoint matrices according to \(\rho(a)=a/5\), with environment-side transition matrices specified in Appendix~\ref{app:generative_process}.

\subsection{Per-Partner Generative Models}
\label{sec:partner_models}

Rather than maintaining a single pooled social model, the focal agent maintains a separate discrete generative model $\mathcal{M}_k$ for each partner $k$. This factorization keeps observations, beliefs, and confidence evidence local to the relationship that generated them. Each $\mathcal{M}_k$ uses the standard $A$/$B$/$C$/$D$/$E$ components of discrete active inference~\cite{DaCosta2020DiscreteSynthesis}: observation likelihoods, transition dynamics, preferences, initial priors, and policy priors. The focal agent's partner-local POMDP specification is provided in Appendix~\ref{app:pomdp}.

Each partner model contains partner type, partner stance, and an own-investment state factor. Type and stance are inferred from evidence; own investment is deterministically set by the agent's executed action and treated as fully observed. This factor lets the payoff likelihood condition on the investment level the agent just chose, so the agent is uncertain about partner type and stance but not about its own investment. Observations are the partner's cooperate-or-defect response and the focal agent's realized payoff. The partner-response likelihood depends on type and stance, so predictive reliability is defined by response model fit rather than payoff: a reliably cooperative and a reliably defecting partner can both be predictable. The payoff likelihood depends on investment level and marginalizes over partner response. The agent's transition model mirrors the process dynamics in Appendix~\ref{app:generative_process}, while scheduled betrayal and repair manipulations are environment-side interventions defined in Appendix~\ref{app:protocols}. This model supplies ordinary partner-state inference; the affective-precision layer in Section~\ref{sec:affective_precision} estimates how confidently each partner model should be deployed during policy selection.

\subsection{Partner-Local Affective Precision}
\label{sec:affective_precision}

Partner-local affective precision maps relationship-specific model fit onto policy precision. At each round \(t\), before assimilating the observed partner response, the focal agent evaluates how surprising that response would be under its predictive type--stance posterior for partner \(k\). After this precision update, the partner-state posterior is updated on the same observation. Unlike prior affective-precision accounts centered on policy-inference quantities~\cite{Hesp2021DeeplyFeltAffect}, the evidence source here is perceptual partner-response model fit. Because this update is computed separately for each partner, confidence can diverge across relationships even when the focal agent's task and preferences remain fixed.

Let \(\hat{o}_{k,t}^{\text{act}}\) denote the observed response from partner \(k\) at round \(t\), \(q_{k,t}^-\) the type--stance posterior before updating on \(\hat{o}_{k,t}^{\text{act}}\), and \(p_{k,t}\) the posterior-predictive probability assigned to that response. The agent computes partner-response likelihood surprisal as
\begin{equation}
\epsilon_{k,t} = -\log p_{k,t} = -\log \sum_{s^{\text{type}},s^{\text{stance}}}
P\!\left(\hat{o}_{k,t}^{\text{act}} \mid s^{\text{type}},s^{\text{stance}}\right)
q_{k,t}^-(s^{\text{type}},s^{\text{stance}}).
\label{eq:epsilon}
\end{equation}
Equation~\eqref{eq:epsilon} is the negative log posterior predictive likelihood assigned to the observed partner response under \(q_{k,t}^-\), marginalizing over the agent's pre-update beliefs about partner type and stance. Partner response directly tests the type-by-stance model: a reliably defecting partner can produce low surprisal despite poor payoff, while an erratic cooperative partner can produce high surprisal despite favorable outcomes. The signal therefore tracks partner-response model fit rather than partner payoff.

Per-round update order is: (i) plan from \(q_{k,t}^-\); (ii) observe \(\hat{o}_{k,t}^{\text{act}}\); (iii) compute \(\epsilon_{k,t}\) and update \(\beta_k\); (iv) assimilate \(\hat{o}_{k,t}^{\text{act}}\) into the partner-state posterior.

The neutral baseline is the surprisal of a uniform binary prediction, \(\sigma_0=-\log 0.5=\log 2\). We define affective charge as
\begin{equation}
\phi_{k,t}
= \alpha\!\left(\sigma_0 - \epsilon_{k,t}\right)
= \alpha \log\frac{p_{k,t}}{0.5},
\label{eq:charge}
\end{equation}
where \(\alpha>0\) controls the gain on this evidence. Thus, \(\phi_{k,t}\) is the scaled log evidence of the focal agent's response prediction relative to a uniform chance-level predictor: it is positive when \(p_{k,t}>0.5\), negative when \(p_{k,t}<0.5\), and zero at \(p_{k,t}=0.5\).

The signed charge $\phi_{k,t}$ is then used as evidence over the partner-specific inverse-precision estimate $\beta_k$. Positive charge shifts posterior mass toward lower $\beta_k$, corresponding to higher policy precision; negative charge shifts posterior mass toward higher $\beta_k$, corresponding to lower policy precision. A persistence prior smooths this update across rounds so that confidence changes gradually rather than resetting after each interaction. The full categorical update is given in Appendix~\ref{app:affective_update}.

The agent then sets partner-specific policy precision from the posterior mean:
\begin{equation}
\begin{aligned}
\bar{\beta}_k &= \sum_{\ell} \beta_\ell \, q(\beta_\ell), \\
\gamma_k &= \gamma_{\text{base}} / \bar{\beta}_k,
\end{aligned}
\label{eq:gamma_update}
\end{equation}
where $\gamma_{\text{base}}$ is the baseline policy precision and $q(\beta_\ell)$ is the categorical mass on inverse-precision level $\beta_\ell$. Because $\beta_k$ is inverse precision, higher $\bar{\beta}_k$ yields more tentative policy commitment, whereas lower $\bar{\beta}_k$ yields higher $\gamma_k$ and sharper policy commitment.

The $\beta_k$ posterior is maintained as an auxiliary precision tracker rather than as a hidden state inside the focal agent's generative model. This keeps the mechanism lightweight and experimentally isolable, but prevents the agent from forming prior expectations over its own future confidence states; this limitation is discussed in Section~\ref{sec:discussion}.

\subsection{Cross-Partner Policy Selection}
\label{sec:cross_partner_policy}

In partner-choice regimes, the agent must decide which partner-specific policy candidate to execute. Each per-partner model \(\mathcal{M}_k\) first evaluates policies \(\pi \in \Pi_k\) directed toward partner \(k\). Let \(u_{k,\pi}\) denote the policy score for candidate \(\pi\) under partner model \(\mathcal{M}_k\) prior to the cross-partner precision transformation; under the uniform policy prior used here, this is the negative expected-free-energy score returned by policy inference, with larger values indicating greater policy support~\cite{Friston2017ProcessTheory,Heins2022Pymdp}. The cross-partner step then combines these partner-indexed candidates into a single choice set. This step does not add a new state-inference model; it specifies how partner-local precision \(\gamma_k\) enters the final comparison among partner--investment candidates.

To apply \(\gamma_k\) locally, we separate each partner's mean policy evidence from the relative differences among that partner's policies. Let \(\bar{u}_k = \frac{1}{|\Pi_k|}\sum_{\pi\in\Pi_k} u_{k,\pi}\) be the mean policy evidence for partner $k$. The precision-adjusted score is then
\begin{equation}
\tilde{u}_{k,\pi} = \bar{u}_k + \gamma_k\left(u_{k,\pi} - \bar{u}_k\right),
\label{eq:centered_policy_scores}
\end{equation}
where $\gamma_k$ is the partner-specific policy precision from Eq.~\eqref{eq:gamma_update}. Partner-specific precision \(\gamma_k\) sharpens or flattens commitment among policies involving partner \(k\), without by itself changing the average evidence for selecting that partner. Higher \(\gamma_k\) amplifies within-partner differences, producing more decisive policy commitment; lower \(\gamma_k\) compresses them, producing more tentative commitment.

The final policy posterior is computed by applying a softmax over the transformed scores \(\tilde{u}_{k,\pi}\) across all partners and candidate policies. This procedure does not change the per-partner generative models; it only determines how partner-local precision enters cross-partner policy comparison. With six investment actions and planning horizon \(H=4\), each partner model evaluates all \(6^4=1{,}296\) candidate policies. In partner-choice conditions with four partners, the combined posterior therefore spans \(5{,}184\) partner--policy candidates. Reported policy entropy is the natural-log Shannon entropy of this posterior, whose maximum is \(\log(5{,}184)\approx8.55\) nats.

\subsection{Simulation Setup}
\label{sec:simulation_setup}

Simulations compare four precision conditions: full partner-local precision, a shared-$\beta$ ablation that pools evidence across partners while keeping per-partner POMDP beliefs intact, a tracked-only ablation that updates $q(\beta_k)$ but fixes $\gamma_k = \gamma_{\text{base}}$, and a no-affect control that disables the $\beta$ tracker entirely. Conditions include open graded investment, partner choice, abrupt betrayal, and precision-dynamics/profile diagnostics. Full condition definitions, episode lengths, and metrics are provided in Appendix~\ref{app:protocols}. Code and experiment configurations are available at~\githubrepourl. Uncertainty estimation procedures are detailed in Appendix~\ref{app:protocols}.
\section{Results}
\label{sec:results}

Results test whether partner-local precision tracks predictability over payoff, acts through the \(\beta_k \rightarrow \gamma_k\) deployment pathway, and remains behaviorally consequential under partner choice, betrayal, and profile variation.

\subsection{Partner-Local Precision Tracks Predictability Rather than Realized Payoff}
\label{sec:predictability_value}

\begin{figure}[t]
\centering
\includegraphics[width=\linewidth]{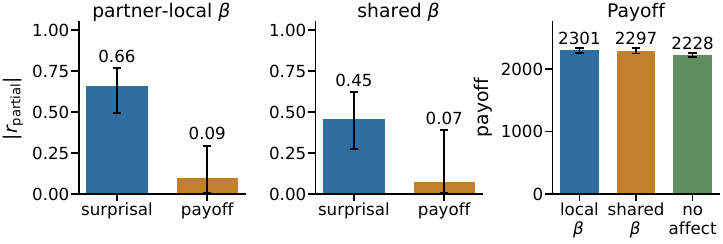}
\Description{Three bar charts compare absolute partial correlations of precision with surprisal and payoff for partner-local and shared precision, followed by payoff for local, shared, and no-affect conditions. Surprisal correlations exceed payoff correlations in both precision conditions. Error bars indicate uncertainty.}
\caption{Partner-local versus shared precision in the locality probe (30 seeds per condition). Left and middle: absolute active-encounter partial correlations of precision with surprisal and payoff, controlling for encounter count and the alternative signal. Right: cumulative payoff. Error bars: 95\% CIs.}
\label{fig:model_fitness_contrast}
\end{figure}

Partner-local precision was more strongly associated with partner-response likelihood surprisal than with realized payoff. Because partner-response surprisal drives the update, we use the full multi-partner simulation to check whether this intended selectivity persists once confidence updating, partner choice, and policy selection interact. Partner-choice episodes couple payoff, surprisal, and exposure, so we computed active-encounter partial correlations controlling for encounter count and the alternative signal. The precision--surprisal association remained substantially stronger ($r=-0.660$, 95\% CI [$-0.766$, $-0.492$]) than the precision--payoff association ($r=0.094$, 95\% CI [$-0.112$, $0.291$]). The negative sign reflects greater $\beta_k$-derived policy precision for less surprising responses; Figure~\ref{fig:model_fitness_contrast} plots absolute magnitudes for readability.

Pooling precision evidence across partners weakened this relationship-specific signal. In the shared-$\beta$ ablation, per-partner beliefs remained separate but the precision tracker was updated from pooled partner evidence. The precision--surprisal association was weaker ($r=-0.454$, 95\% CI [$-0.619$, $-0.275$]), while the precision--payoff association remained small ($r=0.072$, 95\% CI [$-0.163$, $0.390$]). Thus, pooling preserves separate partner beliefs but weakens the alignment between confidence and the relationship-specific predictive history from which that confidence is derived.

Payoff in the same analysis provides behavioral context rather than an independent definition of the mechanism. The partner-local precision signal remains much more strongly associated with partner-response predictability than realized payoff, consistent with confidence in the current partner model rather than reward tracking.

\subsection{Tracked-Only Ablation Isolates Policy Commitment}
\label{sec:deployment}

To test whether partner-local affective precision changes behavior through policy deployment rather than partner-state inference, we used a tracked-only ablation. This variant updates $q(\beta_k)$ normally while holding $\gamma_k=\gamma_{\text{base}}$, preserving the precision estimate but preventing it from modulating policy selection.

\begin{figure}[t]
\centering
\includegraphics[width=\linewidth]{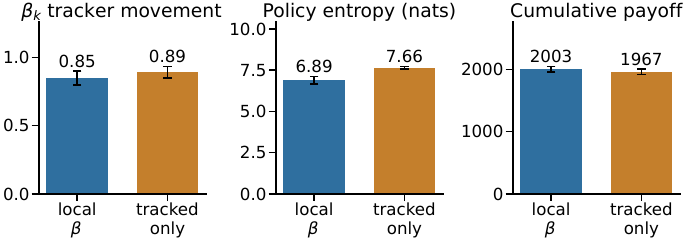}
\Description{Three bar charts compare partner-local and tracked-only conditions. Tracker movement is similar; policy entropy is lower with partner-local precision. The third panel reports cumulative payoff. Error bars indicate uncertainty.}
\caption{Tracked-only ablation in the open graded partner-choice regime. Bars show means across 30 seeds for within-episode $\bar{\beta}_k$ range, policy entropy, and cumulative payoff (95\% CIs). Partner-local precision lowers entropy while tracker movement remains comparable. Paired contrasts are reported in the text.}
\label{fig:deployment_contrast}
\end{figure}

The tracked-only ablation confirmed that the tracker can move without producing the same deployment effect (Figure~\ref{fig:deployment_contrast}). In the open graded partner-choice condition, partner-local and tracked-only conditions showed comparable within-episode variation in posterior mean $\bar{\beta}_k$, differing by only $0.043$ in mean range. The key difference was deployment: full precision modulation lowered policy entropy by $0.770$ nats relative to tracked-only (95\% CI [$-0.966$, $-0.577$]). Cumulative payoff was also somewhat higher under partner-local precision ($2003.3$ versus $1966.8$; difference $=36.5$, 95\% CI [$-8.6$, $84.1$]). Thus, $\beta_k$ matters behaviorally when it is allowed to modulate $\gamma_k$, producing sharper policy commitment. When that pathway is cut, the tracker remains active but the same entropy reduction is lost.

This supports the interpretation of affective precision as a calibration layer rather than an inference-improvement mechanism. The tracker does not directly change the partner-state update; it regulates how strongly the agent commits to policies based on its current partner model. Whether sharper commitment improves payoff depends on the social context, not on precision alone.

\subsection{Partner Choice Extends Policy Commitment across Social Options}
\label{sec:partner_choice}

Using the same underlying partner-choice trajectories analyzed in Section~\ref{sec:deployment}, we next examine a distinct readout: how precision deployment shapes commitment across partner--investment policies and how selected interactions are distributed across partner types. In this regime, each candidate pairs a partner with a multi-step investment plan, of which only the first step is executed each round. Precision modulation lowered policy entropy relative to the no-affect control ($6.887$ versus $7.657$; difference $=-0.770$, 95\% CI [$-0.966$, $-0.577$]), indicating that the $\beta_k \rightarrow \gamma_k$ pathway shapes commitment over partner--investment policies, not only investment levels within a fixed interaction.

Partner selection itself remained broadly distributed across the four partner types (cooperator $29.6\%$ versus $30.7\%$; exploiter $24.4\%$ versus $22.6\%$; reciprocator $22.0\%$ versus $24.0\%$; random $24.0\%$ versus $22.7\%$). The type-specific differences were small, and the paired 95\% bootstrap interval for each selected-type difference included zero. These allocation percentages therefore should not be read as evidence of a general preference for any one partner type. The main effect remains sharper commitment over partner--investment policies rather than systematic selection of a particular social type.

Thus, partner-local precision sharpens policy commitment without imposing a fixed social preference; the abrupt-betrayal condition next tests this mechanism under social change.

\subsection{Abrupt Betrayal Tests Confidence under Change}
\label{sec:betrayal}

Accumulated confidence helps when it remains attached to a valid partner model, but can lag behind abrupt social change. The abrupt-betrayal condition tests this boundary: partner~0 begins as an exploiter in a trusting stance and switches to a hostile stance at round~31, and the agent must adapt while carrying confidence built from the pre-switch relationship.

\begin{figure}[!ht]
\centering
\includegraphics[width=\linewidth]{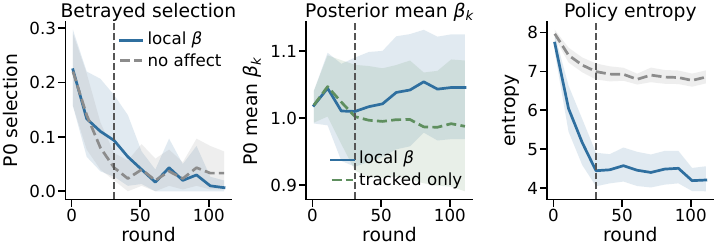}
\Description{Three time-series panels show selection of partner zero, its posterior mean inverse precision, and policy entropy. Dashed vertical lines mark the round-31 stance switch. Solid and dashed condition trajectories have shaded uncertainty bands.}
\caption{Abrupt betrayal in the partner-choice regime. Left and right panels compare partner-local precision with no-affect; the middle panel compares partner-local precision with tracked-only. Dashed vertical lines mark partner~0's round-31 stance switch. Lines show means within 10-round bins across 30 seeds; shading: 95\% CIs.}
\label{fig:betrayal_boundary}
\end{figure}

Precision modulation remained behaviorally consequential through the betrayal episode (Figure~\ref{fig:betrayal_boundary}). Mean policy entropy was lower under partner-local precision than under the no-affect control, with a mean difference of $-2.149$ nats (95\% CI [$-2.379$, $-1.895$]). Joint type--stance accuracy was also higher by $0.158$ (95\% CI [$0.084$, $0.226$]), consistent with altered engagement changing the evidence available to the agent. Because the tracked-only ablation in Section~\ref{sec:deployment} isolated the $\beta_k \rightarrow \gamma_k$ pathway, this accuracy difference is best interpreted as a downstream effect of changed engagement and sampling, not as a direct improvement to the partner-state update.

Cumulative payoff was also higher by $58.2$ (95\% CI [$37.5$, $82.3$]), although this regime-specific difference does not imply that greater precision generally maximizes reward.

The betrayal result exposes a timing problem in social confidence: confidence built from reliable past evidence can remain behaviorally active after the relationship changes. We next vary precision gain and prior model fitness to examine how confidence-revision dynamics differ across agents.

\subsection{Precision Dynamics as Computational Profiles of Social Trust Calibration}
\label{sec:phenotypes}

The betrayal result shows that confidence revision has temporal structure: confidence can be too weak, too slow, or too reactive. Varying precision gain $\alpha$ confirmed that gain controls the amplitude of $\beta_k$ dynamics: in the betrayal regime, the average within-episode max--min range of posterior mean $\bar{\beta}_k$ increased monotonically from $0.097$ at $\alpha=0.05$ to $0.675$ at $\alpha=8.0$ (Spearman $\rho=1.0$). Payoff did not follow this monotonic pattern, indicating that stronger confidence responsiveness is not automatically better calibration.

Crossing prior model fitness with gain produced distinct trust-calibration profiles. Low gain muted confidence dynamics whereas high gain amplified revision, while naive and cautious priors placed the agent at different initial levels of confidence in its partner models. These differences shaped how confidence accumulated and responded to subsequent social evidence. In the tested betrayal profiles, the anxious-reactive configuration (naive prior, high gain) had the highest mean payoff ($2285.8$), but the profiles continued to express distinct confidence-revision dynamics. Exact priors, gain values, and updated profile metrics are reported in Appendix~\ref{app:extended_results}.

Trust-repair diagnostics separately examined reengagement, restored confidence, and payoff recovery after a partner returned to cooperative behavior. These readouts remained dissociable across parameter settings, indicating that renewed interaction need not coincide with identical confidence dynamics. Extended profile and repair metrics are reported in Appendix~\ref{app:extended_results}.
\section{Discussion}
\label{sec:discussion}

The results support partner-local affective precision as a calibration mechanism for social policy commitment. Precision was defined from partner-response predictability rather than payoff, lost its policy-sharpening effect when the $\beta_k \rightarrow \gamma_k$ pathway was cut, and remained behaviorally active under abrupt change without guaranteeing payoff improvement. Together, these findings support the paper's central distinction: social prediction and social commitment are related but distinct. A reliably cooperative partner and a reliably defecting partner can both generate low partner-response likelihood surprisal because both are predictable; conversely, a favorable but inconsistent partner can remain difficult to model. Partner-local affective precision therefore estimates how well the current partner model is working, then regulates how strongly that model is expressed in policy.

The tracked-only ablation isolates the role of precision deployment. Updating $\beta_k$ without allowing it to modulate $\gamma_k$ preserved the tracker but removed the entropy reduction, while full precision modulation sharpened policy commitment. This supports the interpretation that partner-local affective precision does not change the partner-state update rule directly; any accuracy differences that emerge under partner choice are downstream sampling effects, reflecting which partners the agent engages and observes, while the mechanism itself changes how strongly current partner beliefs are deployed. It also clarifies the distinction from affective-precision accounts centered on policy-inference quantities: here, the model-fitness signal is partner-response likelihood surprisal under the pre-update predictive type--stance posterior $q_{k,t}^-$.

The abrupt-betrayal condition highlights an important limitation of the present calibration mechanism. Confidence built from reliable past evidence can remain behaviorally active after the relationship changes. In the present model, $\beta_k$ is an auxiliary tracker, so the agent does not explicitly infer when its own confidence has become stale. Future work could incorporate volatility or change detection that discounts accumulated confidence when prediction-error structure shifts; representing confidence within a deeper generative hierarchy is another possible extension.

The profile analyses generate computational hypotheses about trust calibration, with potential links to work on social learning, psychosis, and attachment~\cite{Adams2013ComputationalPsychosis,Behrens2008SocialLearning,Cittern2018AttachmentActiveInference}. Precision gain and prior model fitness shaped confidence amplitude, adaptation, and repair, but no profile was uniformly best across the reported readouts. Higher gain produced larger confidence swings without monotonic payoff improvement, while gain--prior combinations produced distinct calibration dynamics. These tradeoffs motivate empirical tests of whether human trust-game behavior separates predictability from payoff, policy deployment from partner-state inference, and reengagement from restored confidence. A further extension is to replace scripted partners with reciprocal active-inference agents.

\section{Conclusion}
\label{sec:conclusion}

We formalize social affective precision as a relationship-specific metacognitive signal: an estimate of how confidently an agent should select policies from its current model of each partner. Implemented as partner-local affective precision, this signal updates from partner-response predictive evidence and modulates policy precision during selection. Across simulations in a graded multi-partner trust game, precision tracks predictability more closely than payoff, affects behavior through the $\beta_k \rightarrow \gamma_k$ policy-precision pathway rather than by directly modifying partner-state inference, and remains behaviorally active when relationships change.

These results support a distinction between social prediction and social commitment: an agent can infer what a partner is likely to do while separately estimating how strongly that inference should guide policy commitment. Precision gain and prior model fitness shape this calibration process, producing tradeoffs in confidence accumulation, betrayal adaptation, reengagement, and repair. Future empirical work can test whether human trust-game behavior similarly separates predictability from payoff, inference from policy commitment, and reengagement from restored confidence.

% appendix starts on a fresh page (supplementary; separate from main-text page budget)
\clearpage
\appendix
\renewcommand{\thesection}{\arabic{section}}
% Name appendix headings while keeping numeric cross-references and subsections.
\makeatletter
\newcommand{\appendixprefix@section}{Appendix~}
\renewcommand{\@seccntformat}[1]{\csname appendixprefix@#1\endcsname\csname the#1\endcsname\quad}
\makeatother
% Keep appendix PDF destinations distinct from main-text section numbers.
\renewcommand{\theHsection}{appendix.\arabic{section}}
\raggedbottom
\section{Trust-Game Generative Process}
\label{app:generative_process}
\label{app:supplementary}

\appendixsupplementarystatement

This appendix specifies the environment-side trust-game process: scripted partners with hidden type and stance, investment-dependent stance transitions, binary responses, and deterministic payoff delivery. Appendix~\ref{app:pomdp} specifies the focal agent's POMDP model.

\subsection{Episode Structure}

Each episode consists of repeated interactions between the focal agent and four environment-side partners. On each round, the partner is assigned or selected, the focal agent chooses \(a\in\{0,1,2,3,4,5\}\), the partner emits \(o^{act}_k\in\{\mathrm{cooperate},\mathrm{defect}\}\), payoff is delivered by Eq.~\eqref{eq:graded_payoff_map}, and stance updates by Eq.~\eqref{eq:b_stance}.

\subsection{Partner State Variables}

Each environment-side partner \(k\) has two process variables hidden from the focal agent: behavioral type \(s^{\mathrm{type}}_k \in \{\mathrm{cooperator},\mathrm{reciprocator},\mathrm{exploiter},\mathrm{random}\}\) and stance toward the focal agent \(s^{\mathrm{stance}}_k \in \{\mathrm{trusting},\mathrm{neutral},\mathrm{hostile}\}\). These govern how the environment samples partner responses and updates stance over the episode; type sets the partner's broad response tendency, and stance modulates how cooperative those responses are toward the focal agent at the current interaction.

\subsection{Partner-Response Probabilities}
\label{app:process_response}

The partner cooperation likelihood table defines the probability the partner cooperates given type and stance:

\Needspace{14\baselineskip}
\begin{table}[H]
\centering
\caption{Cooperation likelihood $P(\text{cooperate} \mid s^{\text{type}}, s^{\text{stance}})$}
\label{tab:coop_likelihood}
\setlength{\tabcolsep}{4pt}
\begin{tabular}{lccc}
\toprule
Partner type & Trusting & Neutral & Hostile \\
\midrule
Cooperator & 0.95 & 0.80 & 0.55 \\
Reciprocator & 0.90 & 0.70 & 0.30 \\
Exploiter & 0.70 & 0.35 & 0.10 \\
Random & 0.60 & 0.50 & 0.35 \\
\bottomrule
\end{tabular}
\end{table}

The table defines the environment-side cooperation probability \(P(o^{\mathrm{act}}_k=\mathrm{cooperate}\mid s^{\mathrm{type}}_k,s^{\mathrm{stance}}_k)\), with \(P(o^{\mathrm{act}}_k=\mathrm{defect}\mid \cdot)=1-P(o^{\mathrm{act}}_k=\mathrm{cooperate}\mid \cdot)\).
The same response probabilities are used in the focal agent's partner-response likelihood \(A^{\mathrm{act}}_k\) (Appendix~\ref{app:pomdp_likelihoods}), allowing the focal agent to infer type and stance from observed cooperate-or-defect responses.

\subsection{Payoff Function}
\label{app:process_payoff}

Payoff is delivered by the environment after the partner response is sampled. With endowment \(E=10\) and multiplier \(M=3\), the invested amount is subtracted from the focal agent's endowment. If the partner cooperates, the investment is multiplied by \(M\) and split evenly, so the focal agent receives \(Ma/2\). If the partner defects, the focal agent receives no return from the investment.
\begin{equation}
\pi(a,o_k^{\mathrm{act}})=
\begin{cases}
E-a+\frac{Ma}{2}, & o_k^{\mathrm{act}}=\mathrm{cooperate},\\
E-a, & o_k^{\mathrm{act}}=\mathrm{defect}.
\end{cases}
\label{eq:graded_payoff_map}
\end{equation}
Thus, \(a=0\) is risk-free, while higher investment increases both possible gains and potential loss.

\subsection{Stance Transition Dynamics}
\label{app:process_stance}

Stance transitions depend on the focal agent's graded investment $a$ by interpolating between a high-investment/trust-building endpoint and a low-investment/withholding endpoint:
\begin{equation}
\rho(a)=\frac{a}{|\mathcal{A}|-1}=\frac{a}{5},
\qquad
B^{\mathrm{stance}}(a)=\rho(a)B^{\mathrm{high}}+(1-\rho(a))B^{\mathrm{low}}.
\label{eq:b_stance}
\end{equation}
High investment shifts dynamics toward trust building; low investment or withholding shifts dynamics toward trust damage; recovery from hostility is slower than collapse.

\Needspace{16\baselineskip}
\begin{table}[H]
\centering
\caption{Endpoint stance transition dynamics. The focal agent's graded investment interpolates between $B^{\mathrm{high}}$ and $B^{\mathrm{low}}$ according to Eq.~\eqref{eq:b_stance}.}
\label{tab:stance_transition}
\setlength{\tabcolsep}{4pt}
\begin{tabular}{lccc}
\toprule
\multicolumn{4}{c}{\textbf{High investment / trust-building endpoint }$B^{\mathrm{high}}$} \\
\midrule
To $\setminus$ From & Trusting & Neutral & Hostile \\
\midrule
Trusting & 0.90 & 0.30 & 0.05 \\
Neutral & 0.10 & 0.60 & 0.35 \\
Hostile & 0.00 & 0.10 & 0.60 \\
\midrule
\multicolumn{4}{c}{\textbf{Low investment / withholding endpoint }$B^{\mathrm{low}}$} \\
\midrule
To $\setminus$ From & Trusting & Neutral & Hostile \\
\midrule
Trusting & 0.10 & 0.05 & 0.02 \\
Neutral & 0.50 & 0.35 & 0.18 \\
Hostile & 0.40 & 0.60 & 0.80 \\
\bottomrule
\end{tabular}
\end{table}

These dynamics instantiate the asymmetry of trust: sustained high investment gradually builds trust, while sustained withholding rapidly damages trust; recovery from hostility is slow. In scheduled betrayal and repair protocols, the environment can additionally intervene on a partner's type and stance according to the protocol schedule; these interventions are not known transition events in the focal agent's POMDP.

\subsection{Boundary of the Current Simulation}

All reported simulations use one focal active-inference agent interacting with four scripted partners that maintain hidden type and stance, sample responses by Appendix~\ref{app:process_response}, and update stance by Appendix~\ref{app:process_stance}; they do not infer policies or maintain affective precision.
\section{Focal-Agent POMDP Specification}
\label{app:pomdp}

This appendix specifies the focal agent's partner-local generative model in standard discrete active-inference form. Each partner \(k\) has a POMDP \(\mathcal{M}_k\) with hidden type, stance, and own-investment state factors. The model follows Appendix~\ref{app:generative_process} but is not identical to the environment: payoff is an observation modality marginalizing over partner response; scheduled betrayal and repair are environment-side; affective precision is external to the POMDP.

\subsection{Partner-Local Factorization}

The focal agent maintains a separate model \(M_k\) for each partner, with partner-specific likelihoods, transitions, preferences, priors, and policy priors. Partner choice is handled outside the individual POMDPs by comparing partner--investment policies after each \(M_k\) has produced policy evidence.

\subsection{Hidden State Factors and Observations}

As described in Section~\ref{sec:partner_models}, the focal agent's generative model for each partner $k$ maintains three hidden factors:
\begin{align}
s_k^{\text{type}} &\in \mathcal{S}^{\text{type}}, \label{eq:s_type} \\
s_k^{\text{stance}} &\in \mathcal{S}^{\text{stance}}, \label{eq:s_stance} \\
s^{\mathrm{own}} &\in \mathcal{A}, \nonumber
\end{align}
where
\begin{align*}
\mathcal{S}^{\text{type}} &= \{\mathrm{cooperator},\mathrm{reciprocator},\mathrm{exploiter},\mathrm{random}\}, \\
\mathcal{S}^{\text{stance}} &= \{\mathrm{trusting},\mathrm{neutral},\mathrm{hostile}\}, \\
\mathcal{A} &= \{0,1,2,3,4,5\}.
\end{align*}
\(s^{\mathrm{own}}\) indexes executed investment, is deterministically set by the agent's action, and is treated as fully observed. The joint hidden state per partner spans \(4\times3\times6=72\) configurations.

Observations are partner response and focal payoff:
\begin{align}
o_k^{\text{act}} &\in \mathcal{O}^{\text{act}}, \nonumber \\
o_k^{\text{pay}} &\in \mathcal{O}^{\text{pay}}, \nonumber
\end{align}
where $\mathcal{O}^{\text{act}}=\{\text{cooperate, defect}\}$ and $\mathcal{O}^{\text{pay}}$ comprises eleven discrete payoff levels (Eq.~\eqref{eq:graded_payoff_map}).

\subsection{Observation Likelihoods}
\label{app:pomdp_likelihoods}

\paragraph{Partner-response likelihood \(A^{\mathrm{act}}_k\).}
\(A^{\mathrm{act}}_k\) uses the cooperation probabilities in Table~\ref{tab:coop_likelihood} and is uniform over \(s^{\mathrm{own}}\); investment affects future responses indirectly through stance transitions.

\paragraph{Payoff likelihood \(A^{\mathrm{pay}}_k\).}
In the environment, payoff is deterministic given investment and partner response (Appendix~\ref{app:process_payoff}). In the focal agent's POMDP, payoff is represented as a likelihood over payoff observations conditional on own investment and latent partner state, marginalizing over partner response:
\begin{equation}
\begin{split}
&P(o^{\mathrm{pay}}_k=v \mid s^{\mathrm{own}},s^{\mathrm{type}}_k,s^{\mathrm{stance}}_k) \\
&\quad= \sum_{o^{\mathrm{act}}\in\{\mathrm{cooperate},\mathrm{defect}\}}
\mathbb{I}\!\left[\pi(s^{\mathrm{own}},o^{\mathrm{act}})=v\right]
P(o^{\mathrm{act}}_k=o^{\mathrm{act}} \mid s^{\mathrm{type}}_k,s^{\mathrm{stance}}_k)
\end{split}
\label{eq:a_pay_marginal}
\end{equation}
This makes higher investment beneficial when cooperation is likely and costly when defection is likely, while keeping payoff preferences inside the generative model.

\subsection{Transition Dynamics}

\paragraph{Type drift \(B_k^{\mathrm{type}}\).}
Partner type is uncontrollable and drifts slowly:
\begin{equation}
P(s_k^{\mathrm{type}\prime}=j\mid s_k^{\mathrm{type}}=i)=
\begin{cases}
1-p_{\mathrm{switch}} & j=i,\\
p_{\mathrm{switch}}/(K_{\mathrm{type}}-1) & j\neq i,
\end{cases}
\label{eq:b_type}
\end{equation}
with $K_{\mathrm{type}}=4$ and default $p_{\mathrm{switch}}=0.05$. Scheduled betrayal scenarios set $p_{\mathrm{switch}}=0$.

\paragraph{Stance dynamics \(B_k^{\mathrm{stance}}\).}
\(B_k^{\mathrm{stance}}(a)\) mirrors the interpolation in Eq.~\eqref{eq:b_stance}, using the endpoint matrices in Table~\ref{tab:stance_transition}. Scheduled betrayal and repair intervene on the environment-side process only; the focal agent is not given the intervention times.

\paragraph{Own-investment state factor \(B^{\mathrm{own}}\).}
\(B^{\mathrm{own}}\) deterministically sets own investment to the executed action:
\begin{equation}
P(s^{\mathrm{own}\prime}=a^\prime \mid s^{\mathrm{own}}, a)=\mathbb{I}[a^\prime=a].
\label{eq:b_own}
\end{equation}
Both \(B_k^{\mathrm{stance}}\) and \(B^{\mathrm{own}}\) are indexed by the same investment action \(a\); stance and own investment are separate hidden-state factors, not separate control factors.

\subsection{Preferences, Priors, and Policy Priors}

\paragraph{Observation preferences \(C_k\).}
There is no direct preference over partner responses:
\begin{equation}
C_k^{\mathrm{act}} = [0,\,0] \quad \text{over } \{\mathrm{cooperate},\mathrm{defect}\}.
\label{eq:c_act}
\end{equation}
Payoff preferences ascend over the graded payoff support generated by Eq.~\eqref{eq:graded_payoff_map}:
\begin{equation}
C_k^{\mathrm{pay}} = \log \operatorname{softmax}\!\left(\mathbf{u}/\tau\right),
\quad \mathbf{u}=(5,6,7,8,9,10,10.5,11,11.5,12,12.5),
\label{eq:c_pay}
\end{equation}
where $\tau$ is a temperature parameter controlling preference sharpness. Instrumental motivation therefore enters through payoff observations and multi-step planning rather than direct preferences over partner responses.

\paragraph{Initial-state priors \(D_k\).}
Type, stance, and own-investment priors are
\begin{align}
D_k^{\mathrm{type}} &= \left[\tfrac{1}{4},\tfrac{1}{4},\tfrac{1}{4},\tfrac{1}{4}\right], \label{eq:d_type}\\
D_k^{\mathrm{stance}} &= [0.2,\,0.6,\,0.2], \label{eq:d_stance}\\
D^{\mathrm{own}} &= \mathrm{Unif}(\{0,1,2,3,4,5\}). \label{eq:d_own}
\end{align}

\paragraph{Policy priors \(E_k\).}
Policy priors are uniform over admissible graded investment sequences at the configured planning horizon. In partner-choice settings, partner selection is handled outside the per-partner POMDP by evaluating partner--investment policy candidates with precision-adjusted scores before executing the chosen investment. This cross-partner comparison is specified in Appendix~\ref{app:affective_update}.

\subsection{Runtime and Precision Boundary}
\label{app:pomdp_beta_auxiliary}

The partner-local POMDPs supply ordinary active-inference belief updating and policy evidence for each partner. They do not contain \(\beta_k\) as a hidden state. Instead, \(\beta_k\) is maintained as an auxiliary partner-local precision tracker, updated from partner-response likelihood surprisal and then mapped to policy precision \(\gamma_k\). This design keeps the POMDP state space compact and makes the precision pathway experimentally isolable. The cost is that the focal agent cannot form prior expectations over its own future confidence states. Appendix~\ref{app:affective_update} specifies the auxiliary update and the cross-partner policy-selection procedure.
\section{Affective-Precision Update and Policy Selection}
\label{app:affective_update}

This appendix specifies the auxiliary affective-precision update layered on top of the focal agent's partner-local POMDPs. The update maps partner-response model fit into a categorical posterior over inverse precision \(\beta_k\), then maps the posterior mean to partner-specific policy precision \(\gamma_k\). The same appendix also specifies how \(\gamma_k\) enters partner-choice policy comparison. At each round \(t\), affective charge is computed from Eq.~\eqref{eq:epsilon} using the pre-update predictive posterior \(q_{k,t}^-\) before the partner-state posterior is updated on \(\hat{o}_{k,t}^{\text{act}}\) (Section~\ref{sec:affective_precision}).

\subsection{Categorical Beta Support}

Each partner maintains a categorical posterior $q(\beta_k)$ over discrete inverse-precision levels $\{0.5, 0.67, 1.0, 1.5, 2.0\}$, updated each round from affective charge $\phi_k$. The categorical representation keeps the inverse-precision convention explicit, bounds the precision dynamics, and allows the update to be analyzed as movement over interpretable confidence states.

\subsection{Persistence Prior}

The update proceeds in three steps. First, a persistence prior smooths the previous posterior via a tridiagonal transition matrix:
\begin{equation}
p(\beta_k) \leftarrow T \cdot q_{\text{prev}}(\beta_k),
\label{eq:persistence}
\end{equation}
where $T$ is tridiagonal with persistence parameter 0.8 and reflecting boundaries, encoding the prior expectation that precision evolves slowly.

\subsection{Charge-Based Pseudo-Likelihood}

Second, a charge-based pseudo-likelihood maps positive charge toward high policy precision (low $\beta$) and negative charge toward low policy precision (high $\beta$):
\begin{equation}
\log \ell(\beta_\ell) = \phi_k \cdot \frac{1}{\beta_\ell},
\label{eq:pseudo_lik}
\end{equation}
where $\beta_\ell$ ranges over the five support points. This auxiliary update satisfies the intended monotonicity: positive charge reinforces lower $\beta$, negative charge reinforces higher $\beta$, and the effect scales with effective precision at each support point.

The fixed chance-level baseline, discrete $\beta_k$ support, persistence transition, and charge-based pseudo-likelihood are modeling choices that provide a bounded and interpretable implementation of the confidence tracker. Alternative formulations could learn partner-specific predictive baselines, adapt confidence persistence to inferred volatility, or represent $\beta_k$ continuously.

\subsection{Posterior Update and Policy Precision}

Third, the posterior is normalized:
\begin{equation}
q(\beta_k) \propto \exp(\log \ell(\beta_\ell)) \cdot p(\beta_k).
\label{eq:posterior_update}
\end{equation}
Policy precision is then set from the posterior mean as given in Eq.~\eqref{eq:gamma_update} of the main text. The quantity $\beta_k$ is inverse precision, so higher posterior mean $\beta_k$ lowers policy precision $\gamma_k$, while lower posterior mean $\beta_k$ raises $\gamma_k$. As noted in Appendix~\ref{app:pomdp_beta_auxiliary}, $\beta_k$ is maintained as an auxiliary tracker outside the POMDP.

\subsection{Cross-Partner Policy Selection}

In partner-choice regimes, partner-specific precision sharpens or flattens policy commitment among policies directed toward a given partner while preserving the mean policy evidence associated with that partner. The transformation in Eq.~\eqref{eq:centered_policy_scores} applies $\gamma_k$ to deviations from the partner-specific policy mean, rather than to raw policy scores. The agent samples or selects policies from the combined set of transformed scores across all partners and policies. This lets $\gamma_k$ control commitment among policies involving partner $k$ without changing the average evidence for approaching partner $k$ during cross-partner comparison.

\subsection{Ablation Conditions}

Precision conditions are summarized in Table~\ref{tab:precision_variants} (see also Section~\ref{sec:simulation_setup}).

\Needspace{10\baselineskip}
\begin{table}[H]
\centering
\small
\caption{Precision ablation conditions}
\label{tab:precision_variants}
\begin{tabularx}{\linewidth}{@{}>{\raggedright\arraybackslash}p{0.18\linewidth}X>{\raggedright\arraybackslash}p{0.24\linewidth}@{}}
\toprule
Condition & $\beta$ tracker and $\gamma_k$ & Precision scope \\
\midrule
No-affect & off; $\gamma_k = \gamma_{\text{base}}$ fixed & --- \\
Tracked-only & $q(\beta_k)$ updated; $\gamma_k = \gamma_{\text{base}}$ fixed & per-partner \\
Partner-local & $q(\beta_k)$ updated; $\gamma_k$ from $q(\beta_k)$ & per-partner \\
Shared-$\beta$ & pooled $q(\beta)$ updated; $\gamma_k$ from shared $\beta$ & pooled tracker; separate beliefs \\
\bottomrule
\end{tabularx}
\end{table}
\section{Simulation Protocols and Metrics}
\label{app:protocols}

This appendix summarizes simulation conditions, replication counts, schedules, and metrics.

\subsection{Simulation Conditions}

All reported experiments use one focal active-inference agent with four partners (\(K=4\)), per-partner generative models, graded investment actions, and environment-side partner policies (Appendices~\ref{app:generative_process}--\ref{app:pomdp}). Conditions differ by assignment mode, scheduled partner changes, and precision/profile manipulations within the graded payoff regime. In conditions without explicit partner choice, the interaction partner is assigned according to the configured protocol; in partner-choice conditions, the agent selects the partner before selecting an investment level. Table~\ref{tab:simulation_conditions} summarizes the condition families.

\Needspace{12\baselineskip}
\begin{table}[H]
\centering
\small
\caption{Simulation condition families and purpose}
\label{tab:simulation_conditions}
\begin{tabularx}{\linewidth}{@{}>{\raggedright\arraybackslash}p{0.22\linewidth}X@{}}
\toprule
Condition & Purpose \\
\midrule
Open partner choice & Tracked-only deployment contrast and descriptive partner-type allocation readout from the same trajectories \\
Locality probe & Precision--surprisal vs.\ precision--payoff correlations under local/shared $\beta$ \\
Abrupt betrayal & Episode-level policy entropy, payoff, and joint type--stance accuracy, with post-switch partner-selection, precision, and entropy trajectories \\
Profile variation & $\alpha$/prior manipulations for trust-calibration profiles \\
Forgiveness & Reengagement and $\beta$ recovery after repair \\
\bottomrule
\end{tabularx}
\end{table}

\subsection{Seed Counts and Episode Lengths}

Behavioral comparisons use 30 seeds per condition and profile sweeps use 20 seeds (Table~\ref{tab:protocol_scales}).

\begin{table}[H]
\centering
\small
\caption{Representative episode lengths and replication counts by experiment family}
\label{tab:protocol_scales}
\begin{tabularx}{\linewidth}{@{}>{\raggedright\arraybackslash}p{0.28\linewidth}Xr@{\hspace{8pt}}r@{}}
\toprule
Experiment family & Setting & Rounds & Seeds \\
\midrule
Open deployment / allocation & graded partner choice; shared trajectories for Sections~\ref{sec:deployment}--\ref{sec:partner_choice} & 200 & 30 \\
Locality probe & graded partner choice & 200 & 30 \\
Abrupt betrayal & scheduled partner-choice betrayal & 120 & 30 \\
Profile program & sweeps, factorial, forgiveness & 200 & 20 \\
\bottomrule
\end{tabularx}
\end{table}

\subsection{Betrayal Schedule}

Protocol rounds are numbered from 1; analysis converts the zero-based round indices in the raw output accordingly.

The main abrupt-betrayal protocol switches P0 from trusting to hostile at round 31, retaining its exploiter type; other partners retain their initial configurations and spontaneous type switching is disabled. In the gain-sweep and gain--prior profile betrayal protocols, P0 instead changes from trusting cooperator to hostile exploiter at round 81. Forgiveness uses that same round-81 change, then restores cooperator type and trusting stance at round 121: baseline rounds 1--80, betrayal rounds 81--120, and repair rounds 121--200.

\subsection{Metrics}

Table~\ref{tab:metrics_glossary} defines reported metrics.

\Needspace{14\baselineskip}
\begin{table}[H]
\centering
\small
\caption{Metric glossary}
\label{tab:metrics_glossary}
\begin{tabularx}{\linewidth}{@{}>{\raggedright\arraybackslash}p{0.30\linewidth}>{\raggedright\arraybackslash}X@{}}
\toprule
Metric & Definition \\
\midrule
Total payoff & Cumulative focal payoff over episode/window \\
Policy entropy & Natural-log Shannon entropy of $q(\pi)$, averaged over rounds/seeds \\
Joint type--stance accuracy & Mean simultaneous correctness of inferred partner type and stance \\
Selection share & Fraction of rounds each partner is chosen \\
Selection concentration & Gini coefficient over partner-selection distribution \\
Mean within-episode $\bar{\beta}_k$ range & Average within-episode max--min range of posterior mean $\bar{\beta}_k$, averaged across seeds and episodes \\
Investment churn & Fraction of adjacent rounds with changed investment \\
Early exploiter investment & Rounds 1--30 choosing exploiter with investment $\geq$ half range \\
Trust asymmetry & Withdrawal latency / initial high-investment approach latency \\
Recovery rounds & Rounds after switch until partner-0 rolling selection rate reaches 90\% of its pre-switch rate \\
Locality-probe partial (\(r\)) & Partner--seed partial correlation between $\beta$-derived precision and surprisal/payoff. The precision--surprisal correlation controls for payoff and encounter count; the precision--payoff correlation controls for surprisal and encounter count. \\
\bottomrule
\end{tabularx}
\end{table}

\subsection{Uncertainty Estimates}

Except for Figure~\ref{fig:alpha_sweep_appendix}, reported confidence intervals are two-sided 95\% percentile bootstrap intervals based on 10,000 resamples with bootstrap seed 0. Matched variants are compared by resampling seed-level paired differences. Partial-correlation intervals resample complete simulation-seed clusters, and time-course intervals resample seed-level means within each round bin. Figure~\ref{fig:alpha_sweep_appendix} instead uses normal-approximation intervals, computed as the mean $\pm 1.96$ times its standard error across available runs; its entropy panel pools both environments.
\section{Extended Profile and Trust-Repair Results}
\label{app:extended_results}

This appendix reports gain sweeps, computational trust-calibration profiles, and trust-repair analyses.

\subsection{Alpha Sweep}
\label{sec:phenotypes_alpha}

\begin{figure}[H]
\centering
\appendixfigure{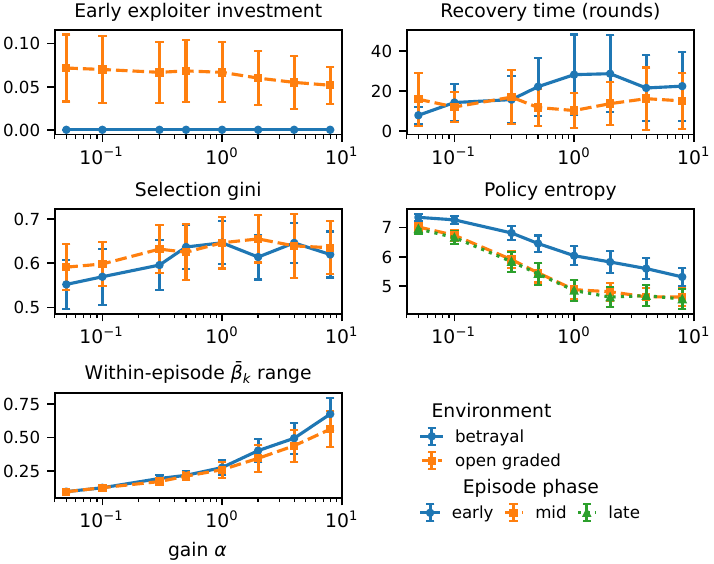}
\Description{Five plots show early exploiter investment, recovery time, selection Gini, policy entropy, and inverse-precision range as precision gain increases on a logarithmic axis. Colors, markers, and line patterns distinguish environments or early, middle, and late episode windows. Inverse-precision range increases with gain.}
\caption{Precision-gain sweep across open graded and betrayal environments (20 seeds per $\alpha$/environment). Points show means with 95\% CIs, excluding undefined recovery times. The entropy panel pools both environments (40 runs per $\alpha$); early, mid, and late denote rounds 1--50, 51--150, and 151--200. All horizontal axes show gain $\alpha$ on a logarithmic scale. Other metric definitions appear in Table~\ref{tab:metrics_glossary}.}
\label{fig:alpha_sweep_appendix}
\end{figure}

Sweeping $\alpha$ across $\{0.05, 0.1, 0.3, 0.5, 1.0, 2.0, 4.0, 8.0\}$ tests whether precision gain controls reactive social confidence across metrics. The average within-episode max--min range of posterior mean $\bar{\beta}_k$ increases monotonically with $\alpha$ in the betrayal regime (Table~\ref{tab:alpha_sweep}). Figure~\ref{fig:alpha_sweep_appendix} shows the sweep across metrics.

\begin{table}[H]
\centering
\caption{Mean within-episode $\bar{\beta}_k$ range by precision-gain $\alpha$ (betrayal regime)}
\label{tab:alpha_sweep}
\begin{tabular}{cc}
\toprule
$\alpha$ & Mean within-episode $\bar{\beta}_k$ range \\
\midrule
0.05 & 0.097 \\
0.1  & 0.126 \\
0.3  & 0.195 \\
0.5  & 0.219 \\
1.0  & 0.277 \\
2.0  & 0.402 \\
4.0  & 0.495 \\
8.0  & 0.675 \\
\bottomrule
\end{tabular}
\end{table}

\subsection{Gain-Prior Profile Signatures}
\label{sec:phenotypes_quadrants}

\begin{figure}[H]
\centering
\appendixfigure{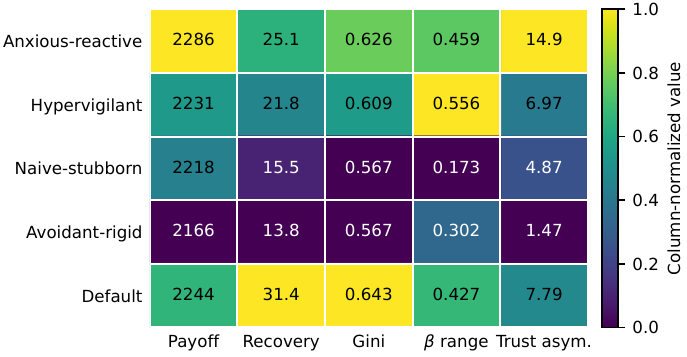}
\Description{A five-by-five heatmap compares computational profiles across payoff, recovery, selection Gini, inverse-precision range, and trust asymmetry. Cell numbers show raw means; color encodes each column normalized separately. A color bar runs from zero to one.}
\caption{Gain-prior profile signatures in the betrayal environment. Printed cell values are raw means; color scales each metric separately from the lowest to the highest profile mean. Thus, colors compare profiles within a column, not magnitudes between metrics. Table~\ref{tab:phenotypes} gives the numerical summary.}
\label{fig:phenotype_quadrants_appendix}
\end{figure}

``Naive'' and ``cautious'' refer to priors over $q(\beta_k)$: naive $[0.40, 0.40, 0.15, 0.04, 0.01]$, cautious $[0.01, 0.04, 0.15, 0.40, 0.40]$. Low/high gain use $\alpha=0.1/3.0$. Profile labels correspond to these combinations: naive-stubborn uses naive prior/low gain, anxious-reactive uses naive prior/high gain, avoidant-rigid uses cautious prior/low gain, and hypervigilant uses cautious prior/high gain. The default uses $\alpha=3.0$ with centered initialization at $\beta_k=1.0$. These labels denote computational parameter combinations rather than clinical classifications or validated human phenotypes. Table~\ref{tab:phenotypes} reports raw metrics and Figure~\ref{fig:phenotype_quadrants_appendix} shows column-normalized profile signatures.

\begin{table}[H]
\centering
\caption{Behavioral metrics by profile in the betrayal environment; the default agent is shown for reference. Values are means from 20 seeds per profile, excluding undefined recovery times. The $\bar{\beta}_k$ range is the within-episode max--min range, averaged over partners and seeds.}
\label{tab:phenotypes}
\small
\begin{tabularx}{\linewidth}{@{}l*{5}{>{\centering\arraybackslash}X}@{}}
\toprule
Profile & Payoff & Recovery (rounds) & Gini & Mean $\bar{\beta}_k$ range & Trust asymmetry \\
\midrule
Anxious-reactive & 2285.8 & 25.1 & 0.626 & 0.459 & 14.87 \\
Hypervigilant & 2230.7 & 21.8 & 0.609 & 0.556 & 6.97 \\
Naive-stubborn & 2218.0 & 15.5 & 0.567 & 0.173 & 4.87 \\
Avoidant-rigid & 2166.0 & 13.8 & 0.567 & 0.302 & 1.47 \\
Default & 2244.2 & 31.4 & 0.643 & 0.427 & 7.79 \\
\bottomrule
\end{tabularx}
\end{table}

\subsection{Forgiveness and Trust Repair}

\begin{figure}[H]
\centering
\appendixfigure{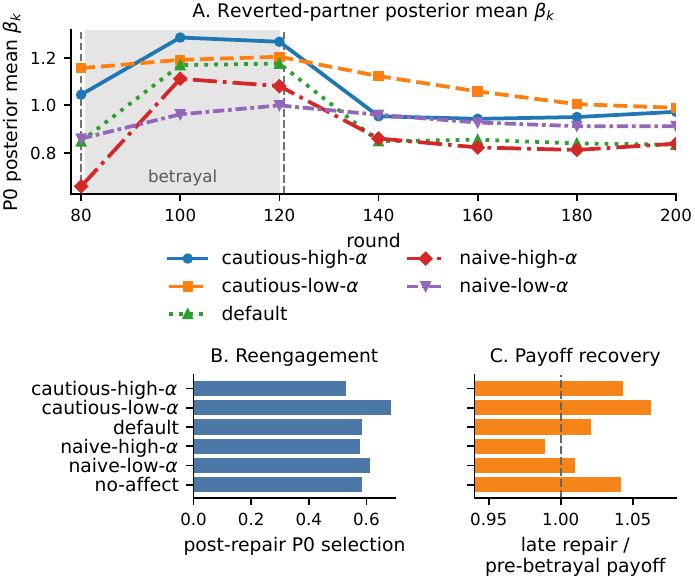}
\Description{Panel A shows partner-zero inverse-precision trajectories for five profiles around betrayal and repair, with a shaded betrayal interval and distinct markers and line patterns. Panels B and C show profile-level reengagement and payoff recovery, including no-affect. A vertical reference line marks a payoff recovery ratio of one.}
\caption{Forgiveness and trust repair (means across 20 seeds per profile). A: partner-0 posterior mean $\beta_k$; shading marks betrayal (rounds 81--120), followed by repair. Lower $\beta_k$ indicates higher policy precision; no-affect has no tracker. B: partner-0 selection during repair. C: payoff in rounds 151--200 relative to rounds 50--80; the dashed line at 1 marks baseline recovery.}
\label{fig:forgiveness_appendix}
\end{figure}

The forgiveness protocol uses baseline rounds 1--80, betrayal rounds 81--120, and repair/reversion rounds 121--200. Table~\ref{tab:forgiveness} reports reengagement, payoff recovery, latency, and mean within-episode $\bar{\beta}_k$ range; Figure~\ref{fig:forgiveness_appendix} shows reengagement, $\beta_k$ recovery, and payoff recovery.

\Needspace{10\baselineskip}
\begin{table}[H]
\centering
\caption{Forgiveness and trust-repair metrics (20 seeds per profile). Reengagement is the post-repair partner-0 selection rate; payoff recovery uses the windows in Figure~\ref{fig:forgiveness_appendix}. Latency averages only seeds that reengage (16--20 per profile). The dash denotes the absent no-affect $\beta$ tracker.}
\label{tab:forgiveness}
\small
\begin{tabularx}{\linewidth}{@{}l*{4}{>{\centering\arraybackslash}X}@{}}
\toprule
Variant & Reengagement & Payoff recovery & Latency (rounds) & Mean $\bar{\beta}_k$ range \\
\midrule
cautious-high-$\alpha$ & 0.529 & 1.043 & 14.5 & 0.607 \\
cautious-low-$\alpha$ & 0.684 & 1.063 & 6.7 & 0.301 \\
default & 0.584 & 1.021 & 10.3 & 0.485 \\
naive-high-$\alpha$ & 0.577 & 0.989 & 3.5 & 0.604 \\
naive-low-$\alpha$ & 0.613 & 1.010 & 7.9 & 0.190 \\
no-affect & 0.586 & 1.042 & 12.4 & -- \\
\bottomrule
\end{tabularx}
\end{table}

\clearpage
\begin{credits}
\subsubsection{\discintname}
The authors have no competing interests to declare that are relevant to the content of this article.
\end{credits}

\bibliographystyle{splncs04}
\bibliography{references}

\end{document}